%% file: template.tex
\documentclass[journal,article,accept, moreauthors,symmetry]{Definitions/mdpi} 
\newcommand{\revision}[1]{#1}
\usepackage{subcaption}

\definecolor{bl}{rgb}{0.25, 0.5, 0.9}
\newcommand{\best}[1]{\textcolor{red}{\textbf{#1}}}   
\newcommand{\second}[1]{\textcolor{blue}{\underline{#1}}}  
\usepackage{multicol, multirow, threeparttable}
\usepackage{booktabs}
\usepackage{tabularx}
\usepackage{subcaption}
\firstpage{1} 
\pubvolume{1}
\issuenum{1}
\articlenumber{0}
\pubyear{2025}
\copyrightyear{2024}
\datereceived{ } 
\daterevised{ } 
\dateaccepted{ } 
\datepublished{ } 
\hreflink{https://doi.org/} 

\Title{TS-HTFA: Advancing Time Series Forecasting via Hierarchical Text-Free Alignment with Large Language Models}

\TitleCitation{Title}

\Author{PengfeiWang $^{1}$\orcidA{}, Huanran Zheng$^{1}$\orcidB{}, Qi'ao Xu$^{1}$\orcidC{}, Silong Dai$^{1}$\orcidD{},
        YiqiaoWang $^{1}$\orcidE{},
        WenJing Yue$^{1}$\orcidF{},
        Wei Zhu$^{1}$\orcidG{},
        Tianwen Qian$^{1}$*\orcidH{} and Xiaoling Wang $^{1}$*\orcidI{}}

\AuthorNames{Firstname Lastname, Firstname Lastname and Firstname Lastname}

\isAPAStyle{%
       \AuthorCitation{Lastname, F., Lastname, F., \& Lastname, F.}
         }{%
        \isChicagoStyle{%
        \AuthorCitation{Lastname, Firstname, Firstname Lastname, and Firstname Lastname.}
        }{
        \AuthorCitation{Lastname, F.; Lastname, F.; Lastname, F.}
        }
}

\address{%
$^{1}$ \quad School of Computer Science and Technology, East China Normal University, Shanghai, 20062, China; \{pfwang,  hrzheng, 52285901023, 51265901134, 51255901045, wjyue, wzhu\}@stu.ecnu.edu.cn, twqian96@gmail.com, xlwang@cs.ecnu.edu.cn}

\corres{Correspondence: xlwang@cs.ecnu.edu.cn;twqian96@gmail.com}

\abstract{Given the significant potential of large language models (LLMs) in sequence modeling, emerging studies have begun applying them to time-series forecasting. Despite notable progress, existing methods still face two critical challenges: 1) their reliance on large amounts of paired text data, limiting the model applicability, and 2) a substantial modality gap between text and time series, leading to insufficient alignment and suboptimal performance. In this paper, we introduce \textbf{H}ierarchical \textbf{T}ext-\textbf{F}ree \textbf{A}lignment (\textbf{TS-HTFA}), a novel method that leverages hierarchical alignment to fully exploit the representation capacity of LLMs while eliminating the dependence on text data. Specifically, we replace paired text data with adaptive virtual text based on QR decomposition word embeddings and learnable prompt. Furthermore, we establish comprehensive cross-modal alignment at three levels: input, feature, and output. Extensive experiments on multiple time-series benchmarks demonstrate that HTFA achieves state-of-the-art performance, significantly improving prediction accuracy and generalization.}


\begin{document}

\input{01_introduction_new}
\input{02_related_work}
\input{03_preliminaries}
\input{04_method}
\input{05_experiment}

\input{06_conclusion}

\funding{This research was funded by NSFC grant (No. 62136002 and 62477014), Ministry of Education Research Joint Fund Project (8091B042239), and Shanghai Trusted Industry Internet Software Collaborative Innovation Center.}

\dataavailability{All the dataset are contained within the article and can be download \url{https://github.com/thuml/Time-Series-Library}.} 

\begin{adjustwidth}{-\extralength}{0cm}

\reftitle{References}
\externalbibliography{yes}
\bibliography{myref} 
\end{adjustwidth}
\end{document}

%% file: 01_introduction_new.tex
\section{Introduction}
Time-series forecasting, as a fundamental task in artificial intelligence research, plays a pivotal role across various domains such as finance \cite{muhammed2024deep}, healthcare \cite{10447194}, environmental monitoring \cite{liu2024wftnet}, and industrial processes \cite{pan2022duma}. It evaluates a model's ability to forecast future trends from historical data, requiring the model to effectively capture intricate temporal dependencies and evolving patterns. With the advancement of deep learning, models like Convolutional Neural Networks (CNNs) \cite{BaiTCN2018, wang2022micn}, Transformers \cite{Wu2022TimesNetT2}, and Multi-Layer Perceptrons (MLPs) \cite{zeng2023dlinear, das2023tide} have been increasingly applied to time-series forecasting, propelling sustained progress in this field.

\begin{figure}[t]
\begin{center}
    \centerline{\includegraphics[width=1\textwidth]{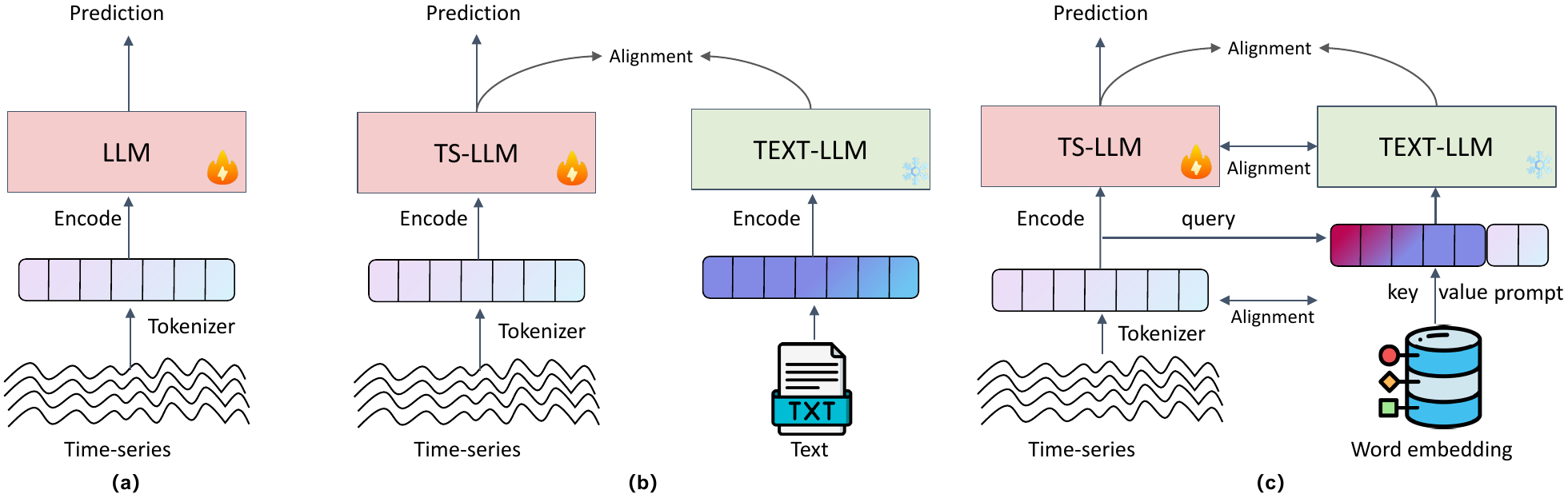}}
    \caption{Comparison on LLMs for Time Series frameworks. (a) Single-Stream Models; (b) Two-Stream Models; (c) TS-HFTA (Ours). }
    \label{fig:motivation}
\end{center}
\end{figure}

Large language models (LLMs) have recently demonstrated exceptional sequence modeling capabilities in language and vision tasks. Leveraging attention mechanisms and large-scale pretraining, LLMs possess powerful semantic representation and can effectively perform reasoning over extended periods. Consequently, they are considered to have substantial potential for addressing time-series forecasting. Several pioneering studies have explored the integration of LLMs, as shown in \cref{fig:motivation} (a) and \cref{fig:motivation} (b), which can be divided into two categories: Single-Stream Models \cite{xue2023promptcast,cao2023tempo,chang2023llm4ts,jin2023time, pan2024textbf, sun2023test} and Two-Stream Models \cite{wang2022open, qiu2023can, li2024frozen, jia2024gpt4mts, yu2024ecg,kim2024eeg}. In single-stream models, time-series signals are initially tokenized and subsequently mapped into the space of large language models through a simple adapter. Then, the model is fully fine-tuned using the time-series data. However, this straightforward approach may cause catastrophic forgetting, leading to degraded model performance. To fully exploit the representational capabilities of LLMs, two-stream models incorporate an LLM branch with frozen parameters that takes paired language descriptions as input. During training, the features from both branches are aligned

Despite notable progress, existing methods still face two critical challenges: 1) Dependence on paired textual annotations. Although incorporating language data can enhance the model performance, time series data, typically collected directly from sensors or smart devices, comes without corresponding language descriptions. Some methods \cite{yu2024ecg,kim2024eeg} attempt to alleviate this limitation by generating synthetic text or relying on small-scale manually labeled data. However, they often fall short regarding sample diversity and temporal consistency with the time-series data. Consequently, the reliance on large-scale high-quality textual annotations restricts the model's scalability in real-world scenarios. 2) Insufficient cross-modal alignment. Time series data is represented as continuous signals, while natural language comprises discrete words, reflecting fundamental differences in their semantic characteristics. Existing methods \cite{li2024frozen, jia2024gpt4mts} commonly project time-series signals and language data into a shared representation space through a single-layer feature mapping and coarse-grained feature alignment. However, they struggle to effectively capture the deep semantic relationships and structural characteristics between the two modalities, leading to suboptimal results.

To address the above-mentioned challenges, we propose a novel method termed as Hierarchical Text-Free Alignment \textbf{(HTFA)}, featuring two primary innovations. Firstly, to overcome the reliance on textual annotations, we hypothesize that the vocabulary of LLMs naturally contains some tokens that are implicitly related to time-series data. Based on this insight, we design a time-series-guided dynamic adaptive gating mechanism that maps time-series data into the feature space of the language model's word embedding, generating corresponding virtual text tokens. Additionally, to avoid the situation where certain specialized signals lack corresponding word embeddings in the existing vocabulary, we introduce prompt tuning, which allows the model to learn new virtual word embeddings. Secondly, to tackle the insufficient cross-modal alignment between time series and language, our theoretical analysis in Section \ref{subsub:Theoretical} concludes that the comprehensive alignment between two modalities necessitates alignment across multi-levels: input distribution, intermediate feature distribution, and output distribution. Based on this theory, we propose a hierarchical alignment strategy. Specifically, at the input level, we apply the dynamic adaptive gating mechanism to map time-series data to virtual text tokens, ensuring consistency between the input encoding. For the intermediate feature, we employ contrastive learning to optimize the representations between the time-series and the language data by minimizing the semantic distance between positive samples. At the output level, we introduce optimal transportation to further align the two modalities in the prediction space. 

The main contributions of this paper are as follows:
\begin{itemize}
    \item We propose a hierarchical text-free alignment with a novel probabilistic analysis, enabling comprehensive alignment across the input, intermediate, and output spaces.
    \item We integrate dynamic adaptive gating in the input embedding layer to align time-series data with textual representations and generate virtual text embeddings.
    \item We introduce a novel combination of layer-wise contrastive learning and optimal transport loss to align intermediate representations and output distributions.
    \item We validate the proposed framework through extensive experiments, achieving competitive or state-of-the-art performance on both long-term and short-term forecasting tasks across multiple benchmark datasets.
\end{itemize}

%% file: 02_related_work.tex
\section{Related Work}
\subsection{Time Series Forecasting}
In recent years, deep learning has revolutionized time series forecasting, with numerous methods emerging to enhance predictive accuracy. Models such as Recurrent Neural Networks (RNNs), Convolutional Neural Networks (CNNs), Multilayer Perceptrons (MLPs), and Transformers have been widely applied to time series analysis. TimesNet \cite{Wu2022TimesNetT2}, a CNN-based model, leverages Fourier Transform to identify multiple periods in time series, converting 1D data into 2D tensors to capture both intra- and inter-period variations using convolutional layers. DLinear \cite{zeng2023dlinear}, an MLP-based model, decomposes raw time series into trend and seasonal components via moving averages, extracts features through linear layers, and combines them for predictions. Transformer-based models have gained significant traction, leveraging their success in NLP. FEDformer \cite{zhou2022fedformer} transforms time series into the frequency domain to compute attention scores and capture detailed structures. PatchTST \cite{nie2022pathtst} segments multivariate time-series channels into patches and processes them with a shared Transformer encoder, preserving local semantics while reducing computational costs and handling longer sequences. Crossformer \cite{zhang2022crossformer} extracts channel-wise segmented representations and employs a two-stage attention mechanism to capture cross-time and cross-dimension dependencies. iTransformer \cite{liu2023itransformer} transposes embeddings from the variable to the temporal dimension, treating each channel as a token and using attention mechanisms to capture inter-variable relationships, achieving excellent performance in most cases. Despite these advancements, challenges remain, such as limited training data, overfitting in specific domains, and the complexity of architectural designs.

\subsection{LLMs for Time Series Forecasting}
The realm of large language models  has experienced remarkable evolution, showcasing exceptional capabilities in natural language processing and finding applications across diverse domains. Recently, efforts have been made to adapt LLMs for time-series analysis to enhance predictive performance. PromptCast \cite{xue2023promptcast} introduces a "codeless" approach to time-series forecasting, moving away from complex architectures. TEMPO \cite{cao2023tempo} focuses exclusively on time-series forecasting by integrating elements like time-series decomposition and soft prompts. LLM4TS \cite{chang2023llm4ts} proposes a two-stage fine-tuning framework to address challenges in incorporating LLMs with time-series data. Time-LLM \cite{jin2023time} reprograms time series using natural language-based prompting and source data modality to leverage LLMs as efficient time-series predictors. CALF \cite{liu2024taming} develops a cross-modal match module to decrease modality gaps, driven by the resemblance between forecasting future time points and the generative, self-regressive nature of LLMs. GPT4TS \cite{zhou2023one} presents a unified framework for many time-series tasks that partially freezes LLMs while fine-tuning specific layers, but acknowledges the substantial data and computational demands of this approach. However, alignment strategies for integrating LLMs with time-series data remain insufficiently effective.

\subsection{Cross-Modal Learning}
Cross-modal fine-tuning enables the transfer of knowledge from models pre-trained on data-rich modalities to data-scarce ones, addressing challenges such as data insufficiency and poor generalization \cite{Shen2023CrossModalFA}. Model reprogramming offers a resource-efficient alternative by repurposing pre-trained models from a source domain to solve tasks in a target domain without fine-tuning, even when the domains differ significantly \cite{chen2024model}. Recent studies have explored transferring the capabilities of large language models (LLMs) to other domains, including vision \cite{pangfrozen, lai2024residual}, audio \cite{Jin2023CrossModalDF}, biology \cite{Vinod2023ReprogrammingPL}, and recommender systems \cite{zhao2024recommender}, providing evidence of the cross-modal transfer capacity of pre-trained models. 
In the context of time series, most research leverages the contextual modeling power of LLMs through fine-tuning to enhance forecasting performance. However, these approaches often overlook the inherent distributional differences between language and time series modalities. In this work, we address this gap by employing three stages alignments to effectively transfer pre-trained language model knowledge to the time series domain.

%% file: 03_preliminaries.tex
\section{Preliminaries}
\subsection{Theoretical Analysis}\label{subsub:Theoretical}
In this study, we employ a dual-tower architecture and triplet-level cross-modal distillation to facilitate the transfer of knowledge from the textual domain (\(\mathcal{D}_{\text{text}}\)) to the time-series domain (\(\mathcal{D}_{\text{time}}\)) via a transformation function \(T\):
\[
T: \mathcal{D}_{\text{text}} \to \mathcal{D}_{\text{time}}
\]

We define the domains from a probabilistic perspective as follows:
\[
\mathcal{D} = \{ p(\mathbf{X}), p(\mathbf{X}, \mathbf{y}), p(\mathbf{y}) \}.
\]

To align the two domains, the goal is to align the input distributions \(p(\mathbf{X}_\text{time})\) and \(p(\mathbf{X}_\text{text})\), the joint distributions \(p(\mathbf{X}_{\text{time}}, \mathbf{y}_{\text{time}})\) and \(p(\mathbf{X}_{\text{text}}, \mathbf{y}_{\text{text}})\), as well as the marginal distributions \(p(\mathbf{y}_{\text{time}})\) and \(p(\mathbf{y}_{\text{text}})\).

Introducing the intermediate layer representation \(h\), we decompose the joint and marginal distributions as follows:
\[
p(\mathbf{X}, \mathbf{y}) = \int p(\mathbf{y} \mid \mathbf{h}) \, p(\mathbf{h} \mid \mathbf{X}) \, p(\mathbf{X}) \, d\mathbf{h},
\]
\[
p(\mathbf{y}) = \int \int p(\mathbf{y} \mid \mathbf{h}) \, p(\mathbf{h} \mid \mathbf{X}) \, p(\mathbf{X}) \, d\mathbf{h} \, d\mathbf{X}.
\]

\textbf{Thus, the goal is to align the two domains by reconciling the input distributions \(p(\mathbf{X})\), the intermediate representations \(p(\mathbf{h} \mid \mathbf{X})\), and the output distributions \(p(\mathbf{y} \mid \mathbf{h})\)}. This alignment ensures that:
\[
p(\mathbf{X}_{\text{time}}, \mathbf{y}_{\text{time}}) \rightarrow p(\mathbf{X}_{\text{text}}, \mathbf{y}_{\text{text}})
\]
and
\[
p(\mathbf{y}_{\text{time}}) \rightarrow p(\mathbf{y}_{\text{text}}).
\]

\textbf{Triplet-Level Alignment Framework:} TS-HFTA integrates triplet-level alignment strategies into a unified framework. By aligning the input, intermediate, and output distributions as analyzed above, TS-HFTA ensures the effective transfer of knowledge from the textual domain to the time-series domain while maintaining both semantic and distributional coherence. The dual-tower architecture fixes the textual branch and adapts the time-series branch, ensuring that knowledge distillation avoids forgetting.

\subsection{Task Definition}



Let \([ \mathbf{X}_1, \mathbf{X}_2, \cdots, \mathbf{X}_T ] \in \mathbb{R}^{N \times T}\) represent a regularly sampled multivariate time-series dataset with \(N\) series and \(T\) timestamps, where \(\mathbf{X}_t \in \mathbb{R}^N\) denotes the values of \(N\) distinct series at timestamp \(t\). We define the model input as a look-back window of length \(L\) at timestamp \(t\), namely $\mathbf{X}_t = [{X}_{t-L+1}, {X}_{t-L+2}, \cdots, {X}_{t} ] \in \mathbb{R}^{N \times L}$. The prediction target is defined as a horizon window of length \(\tau\) at timestamp \(t\), denoted as $\mathbf{Y}_t = [{X}_{t+1}, {X}_{t+2}, \cdots, {X}_{t+\tau} ] \in \mathbb{R}^{N \times \tau}$. The goal of time-series forecasting is to use historical observations \(\mathbf{X}_t\) to predict future values \(\hat{\mathbf{Y}}_t\). A typical forecasting model \(f_\theta\), parameterized by \(\theta\), generates predictions by $\hat{\mathbf{Y}}_t = f_\theta (\mathbf{X}_t)$.

%% file: 04_method.tex
\section{Method Overview}
\begin{figure*}[ht]
    \centering
\includegraphics[scale=0.58]{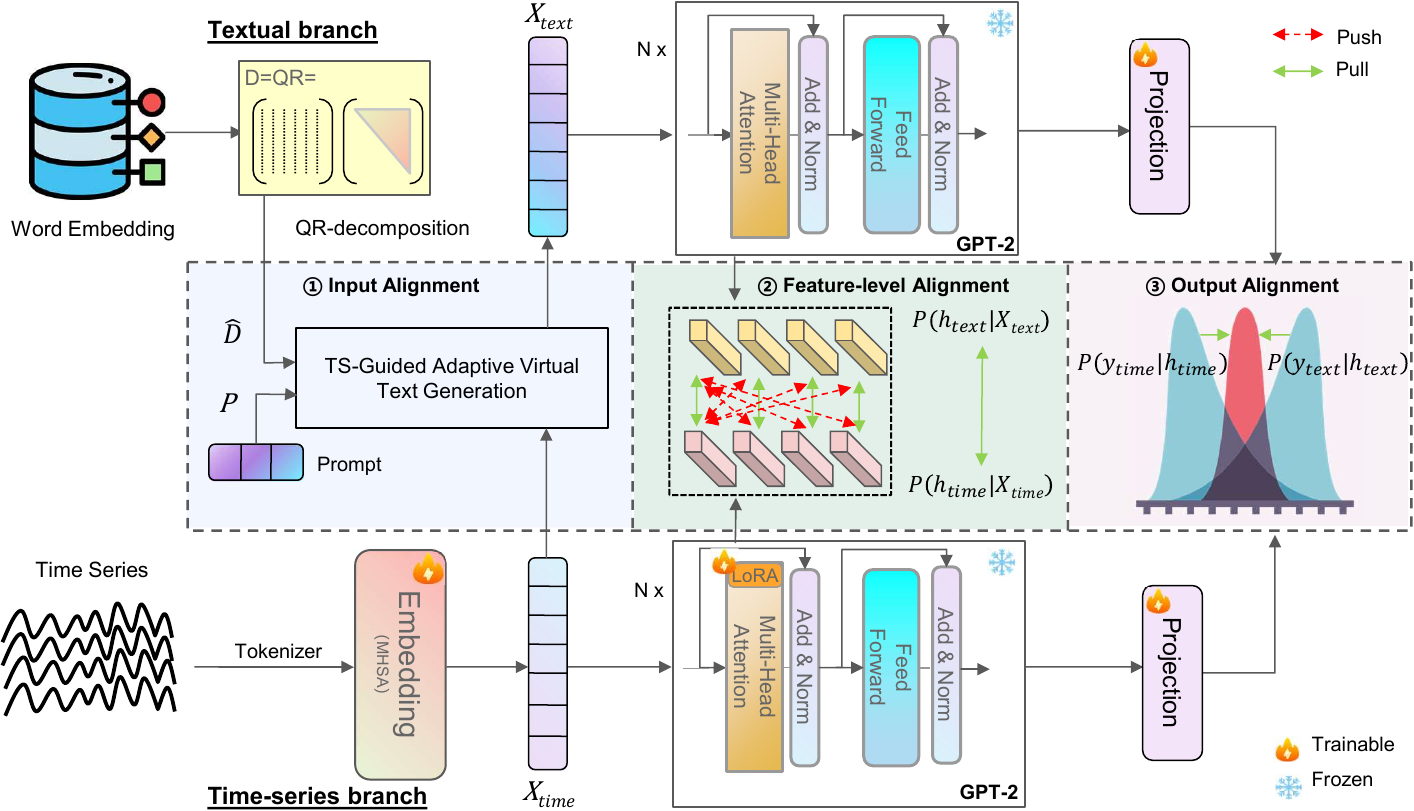}
    \caption{Overview of the proposed \textbf{TS-HFTA}. The framework adopts a dual-branch structure, where the time series branch is hierarchically aligned with the fixed language branch across input, intermediate, and output distributions. During inference, the language branch is excluded, and predictions are generated solely from the time-series branch.}
    \label{fig:framework}
\end{figure*}

As shown in \cref{fig:framework}, the proposed framework adopts a dual-branch structure comprising a time-series branch and a fixed language branch. During the training stage, the time-series branch is trained online using LoRA \cite{hu2021lora} with hierarchical text-free alignment to the language branch across multiple levels: input distribution \cref{subsec:input}, intermediate feature distribution \cref{sub:feature}, and output distribution \cref{sub:ot}. At the input stage, TS-Guided Adaptive Virtual Text Generation (TS-GAVTG) generates a paired virtual text dataset to activate the language branch’s capabilities. Contrastive learning is employed for intermediate feature alignment, while optimal transport (OT) is applied at the output layer to ensure distributional consistency. During inference, the language branch is excluded, and predictions are generated solely by the time-series branch.

\subsection{TS-Guided Adaptive Virtual Text Generation in \textbf{Input} Alignment}\label{subsec:input}
\begin{figure*}[ht]
    \centering
\includegraphics[height=0.25\textheight]{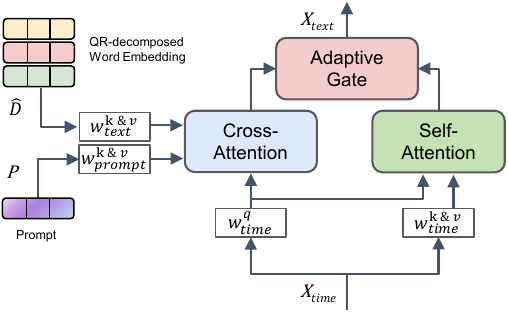}
   \caption{Overview of the proposed \textbf{TS-GAVTG} module for generating virtual paired text data. Time-series data is encoded into tokens and serves as the query in the cross-attention mechanism, with keys and values derived from QR-decomposed word embeddings and learnable prompts. Simultaneously, the self-attention mechanism processes the time series tokens independently. The gating mechanism fuses the outputs from both cross-attention and self-attention to generate virtual paired text tokens.}
    \label{fig:gate}
\end{figure*}

In the input phase, we encode time-series data into tokens, apply QR decomposition to the word embeddings, and generate virtual text tokens using a dynamic adaptive gating mechanism, which is shown in \cref{fig:gate}.
In the input phase, the TS-GAVTG module enhances time-series representation by integrating language priors through a dual-attention mechanism. Time-series data is encoded into tokens that serve as the query in the cross-attention mechanism, while the keys and values are generated from QR-decomposed word embeddings and learnable prompts. The self-attention mechanism processes the time-series tokens independently. A dynamic adaptive gating mechanism fuses the outputs from cross-attention and self-attention to generate virtual paired text tokens, enriching the model's understanding of temporal features.

\subsubsection{Time Series Encoding}\label{subsub:ts}
Given a multivariate time series \( I \in \mathbb{R}^{L \times P} \), where \( L \) is the sequence length and \( P \) is the number of variables, an embedding layer maps each variable across timestamps into a shared latent space\cite{liu2023itransformer}. Multi-Head Self-Attention (MHSA) is then applied to obtain the time series tokens \( X_{\text{time}} = \text{MHSA}(\text{Embedding}(I)) \in \mathbb{R}^{P \times M} \), with \( M \) as the feature dimension of the pre-trained large language models (LLMs). The resulting token set is \( X_{\text{time}} = \{x_i^{\text{time}} \mid i = 1, 2, \ldots, P\} \), where \( x_i^{\text{time}} \) are ordered tokens, and \( i \) denotes their position in the sequence.

\subsubsection{QR-decomposed Word Embedding}\label{subsub:qr} 
In pre-trained LLMs, word embeddings serve as pivotal anchors, structuring and expanding the input distribution within the feature space. To leverage this structured distribution, the vocabulary of the language model \( \hat{D} = \{v_1^{\text{text}}, v_2^{\text{text}}, \ldots, v_Q^{\text{text}}\} \) is derived by performing QR decomposition on the word embedding dictionary \( D \in \mathbb{R}^{|A| \times M} \) to reduce its dimensionality. Specifically, the QR decomposition factorizes the matrix \( D \) into an orthogonal matrix \( Q \) and an upper triangular matrix \( R \). By selecting the first \( d \) vectors from \( Q \) (corresponding to the rank of \( R \)), we construct a reduced principal embedding matrix \( \hat{D} = Q_{[:, :d]} \in \mathbb{R}^{d \times M} \), where \( d \ll |A| \). The orthogonality of \( Q \) ensures that the selected vectors are linearly independent, unlike those obtained through PCA \cite{liu2024taming} or representative selection \cite{sun2023test}, allowing them to effectively span a subspace of the language model’s feature space.

\subsubsection{Virtual Text Generation via Dynamic Adaptive Gating}\label{subsub:dag}

The text representation \( X_{text} \in \mathbb{R}^{P \times M} \) is generated from \( X_{time} \) by applying self-attention within the time series token and prompt-enhanced cross-attention with $\hat{D}$. The attention mechanism is described by:
\begin{equation}
\begin{aligned}
    X_{att} &= {\rm Softmax}\left(\frac{QK^T}{\sqrt{C}}\right)V,  \\
    Q &= X_{time} W_q, \\
    \mathbf{K}_{\text{self}}, \mathbf{V}_{\text{self}} &= X_{time} \mathbf{W}_k^{\text{self}}, X_{time} \mathbf{W}_v^{\text{self}} \text{ (Self-Attention)}, \\
    \mathbf{K}_{\text{cross}}, \mathbf{V}_{\text{cross}} &= [\mathbf{P}_k; \hat{\mathbf{D}} \mathbf{W}_k^{\text{cross}}], [\mathbf{P}_v; \hat{\mathbf{D}} \mathbf{W}_v^{\text{cross}}] \text{ (Cross-Attention)}, \\
\end{aligned}
\end{equation}

The primary difference between self-attention and cross-attention lies in the source of the keys (\(\mathbf{K}\)) and values (\(\mathbf{V}\)), resulting in the obtained \(\mathbf{X}_{\text{self}}\) and \(\mathbf{X}_{\text{cross}}\). \( P_k \in \mathbb{R}^{L_p \times d_k} \) and \( P_v \in \mathbb{R}^{L_p \times d_k} \) are the learnable prompt matrices with \( L_p \) being the prompt length and \( d_k \) being the dimensionality of each attention head.

The prompts guide the cross-attention process by focusing on task-relevant features \cite{lee2023multimodal}, aligning time-series data with text representations, and enhancing the language model's ability to handle unseen time-series tokens. Unlike previous methods \cite{jin2023time, liu2024taming} that assume relevant keys and values can always be retrieved from the language model's vocabulary, our prompt design addresses cases where such key-value pairs are unavailable. 

The final text representation \( X_{text} \) is obtained by adaptively combining the outputs from self-attention and cross-attention:
\begin{equation}
\begin{aligned}
    g = \text{Sigmoid}\left(W_{\text{gate}}^T [X_{cross}; X_{self}]\right)\\
    X_{text} = X_{cross} \odot g + X_{self} \odot (1 - g),
\end{aligned}
\end{equation}

Here \( W_{\text{gate}}^T \in \mathbb{R}^{ 2 d_k \times d_k} \) and \( g\in \mathbb{R}^{d_k \times d_k} \). This dual attention effectively integrates the time series and language information, enabling the generation of virtual text data based on time-series features, thereby facilitating cross-modal learning and downstream applications. The design satisfies the condition that the two spaces of the text and time series branches are not entirely disjoint, as GPT-2 can directly handle initial values or treat them as strings \cite{xue2023promptcast_tkde}. Our gating mechanism is specifically designed to incorporate semantics from compressed word embeddings while retaining numerical values from the time series data, thereby maximizing the potential of the text branch.
\subsection{Layer-Wise Contrastive Learning for \textbf{Intermediate Network} Alignment}\label{sub:feature}
To align the outputs of each intermediate layer in the temporal student branch with those in the textual teacher branch, we use a layer-wise contrastive learning method.
This approach enhances the alignment of semantically congruent information between modalities by optimizing relative similarity in the feature space, rather than minimizing absolute differences \cite{liu2024taming}. Specifically, we apply the InfoNCE loss at each layer:
\begin{equation}
\mathcal{L}_{\text{InfoNCE}}^m = -\frac{1}{N} \sum_{i=1}^{N} \log \frac{\exp\left(\text{sim} \left( F_{\text{time}}^i, F_{\text{text}}^i \right) / \tau \right)}{\sum_{j=1}^{K} \exp\left(\text{sim} \left( F_{\text{time}}^i, F_{\text{text}}^j \right) / \tau \right)},
\end{equation}

where \( \text{sim}(\cdot, \cdot) \) denotes the similarity function, and \( \tau \) is the temperature parameter. The terms \( F_{\text{time}}^i(\cdot) \) and \( F_{\text{text}}^i(\cdot) \) represent the aggregated intermediate features from the temporal and textual modalities, respectively, where average pooling has been applied to combine all tokens within each layer. The overall feature loss \( \mathcal{L}_{\text{feature}} \) integrates these losses across layers:

\begin{equation}
\mathcal{L}_{\text{feature}} = \sum_{m=1}^{L} \gamma^{(L-m)} \mathcal{L}_{\text{InfoNCE}}^m,
\end{equation}
where a decay factor \( \gamma^{(L-m)} \) is applied to give more weight to losses from layers closer to the input, reflecting their importance in the alignment process.

\subsection{Optimal Transport-Driven \textbf{Output Layer} Alignment}\label{sub:ot}

To achieve coherent and unified predictions from both time-series and language modalities, we adopt optimal transport (OT) \cite{villani2021topics} for distributional alignment in the output layer. Unlike traditional \(L_2\) loss, which captures pointwise differences, OT aligns the distributions holistically, effectively preserving semantic relationships across modalities.

The OT loss function is defined as:
\begin{equation}
    \mathcal{L}_{\text{ot}} = \sum_{i,j} P_{ij} \, W_{ij} + \mu H(P),
\end{equation}
where \(P_{ij}\) is the transport plan, \(W_{ij}\) is the cost matrix derived from pairwise distances between \(Y_{\text{time}}\) and \(Y_{\text{text}}\), and  \( H(P) = \sum_{i,j} P_{ij} \log P_{ij} \) is the entropy regularization term controlled by \(\mu\).

We solve this optimization using the Sinkhorn algorithm \cite{Cuturi_2013}, which efficiently computes the transport plan by iteratively normalizing \(P\) under marginal constraints. This approach balances transport cost and regularization, ensuring smooth and stable alignment.

By leveraging OT, our method captures global semantic relationships between modalities, ensuring coherent alignment that goes beyond pointwise matching. Additionally, the use of entropy regularization in the optimization process enables stable and efficient alignment, making the method both robust and computationally efficient.

\subsection{Total Loss Function for Cross-Modal Distillation}\label{sub:all}
To ensure effective training and alignment of the model across modalities, we employ a combined loss function that integrates various components. Specifically, the total loss for the textural student branch, following \cite{liu2024autotimes, liu2024taming}, is a weighted sum of the supervised loss of task \( \mathcal{L}_{\text{task}} = \lvert {\mathbf{Y}}_t - \hat{\mathbf{Y}}_t \rvert \), the feature alignment loss \( \mathcal{L}_{\text{feature}} \), and the output consistency loss \( \mathcal{L}_{\text{ot}} \). The total loss \( \mathcal{L}_{\text{total}} \) is given by:

\begin{equation}
\mathcal{L}_{\text{total}} = \mathcal{L}_{\text{task}} + \alpha \mathcal{L}_{\text{feature}} + \beta \mathcal{L}_{\text{ot}}
\end{equation}

\noindent where \( \alpha \) and \( \beta \) are hyperparameters that balance the contributions.

%% file: 05_experiment.tex
\section{Experiments}
We evaluated our proposed method on standard time-series forecasting benchmarks, comparing it with several state-of-the-art methods for both long-term (as shown in \cref{tab::long-term}) and short-term forecasting (as shown in \cref{tab::short-term}). These comparisons were carried out in line with competitive research \cite{liu2024taming, jin2023time, Wu2022TimesNetT2}. Additionally, ablation studies (as shown in \cref{tab::ablation}) were conducted to assess the impact of each component. All experiments were performed under uniform settings across methods, adhering to established protocols \cite{Wu2022TimesNetT2}, with evaluation pipelines available online\footnote{\url{https://github.com/thuml/Time-Series-Library}}.

\subsection{Experimental Setups}  
\subsubsection{Implementation Details}  
Following \cite{zhou2023one}, we used GPT-2 \cite{radford2019gpt2} with its first six Transformer layers as the backbone. The model was fine-tuned using the LoRA method \cite{hu2021lora}, with a rank of 8, an alpha value of 32, and a dropout rate of 0.1. These values align with commonly adopted settings shown to provide a good balance between performance and parameter efficiency in prior works \cite{liu2024taming}. Similarly, the prompt length was set to 8 based on established configurations \cite{lee2023multimodal} to maintain model flexibility without increasing computational overhead. The Adam optimizer \cite{kingma2014adam} was applied with a learning rate of $5 \times 10^{-4}$. The loss hyperparameters were configured as $\alpha=0.1$ and $\beta=0.01$.  
For the task loss, we adopted smooth $L1$ for long-term forecasting and SMAPE for short-term forecasting. The Optimal Transport parameter was set to $\mu=0.1$, and the number of iterations was fixed at 100. InfoNCE was implemented using the Lightly toolbox \footnote{\url{https://github.com/lightly-ai/lightly}}. To balance the contributions from lower and higher layers, the decay factor was set to $\gamma=0.8$.

\subsubsection{Baselines} 
 We carefully select representative baselines from the current time series forecasting domain. These baselines are categorized as follows: 
 (1) LLMs-based models: CALF \cite{liu2024taming}, TimeLLM \cite{jin2023time} and GPT4TS \citep{zhou2023one}; 
 (2) Transformer-based models: PatchTST~\cite{nie2022pathtst}, iTransformer~\cite{liu2023itransformer}, Crossformer~\cite{zhang2022crossformer}, FEDformer~\cite{zhou2022fedformer}, PAttn \cite{tan2024language} and LTrsf \cite{tan2024language}; 
 (3) CNN-based models: MICN~\cite{wang2022micn} and TimesNet~\cite{Wu2022TimesNetT2}; 
 (4) MLP-based models: DLinear~\cite{zeng2023dlinear} and TiDE~\cite{das2023tide}. 
 Besides, N-HiTS \cite{challu2022nhits} and N-BEATS \cite{oreshkin2019nbeats} are included for short-term forecasting.

\input{05_1_baselines}

\subsubsection{Datasets}
We conduct experiments on two types of real-world datasets, namely long-term forecasting and short-term forecasting.
For long-term Forecasting, we use seven widely-used datasets, including the Electricity Transformer Temperature (ETT) dataset with its four subsets (ETTh1, ETTh2, ETTm1, ETTm2), Weather, Electricity, and Traffic \cite{wu2021autoformer}.
For short-term Forecasting, we adopt the M4 datasets \cite{M4team2018dataset}, which include univariate marketing data collected yearly, quarterly, and monthly. Dataset statistics are summarized in \cref{tab:dataset}.
\input{05_3_dataset}

\subsubsection{Evaluation Metrics}
For both long-term forecasting and short-term forecasting, we provide detailed descriptions of datasets in the corresponding sections. For long-term forecasting, we employ two common evaluation metrics: Mean Square Error (MSE) and Mean Absolute Error (MAE). For short-term forecasting, we apply three evaluation metrics: Symmetric Mean Absolute Percentage Error (SMAPE), Mean Absolute Scaled Error (MASE), and Overall Weighted Average (OWA). The OWA is a specific metric used in the M4 competition.
\input{05_2_metrics}

\textbf{Input Lengths and Prediction Horizons.}
\textit{Long-term Forecasting}: The input time series length $T$ is fixed as $96$ for a fair comparison across datasets. We adopt four distinct prediction horizons $H \in \{96, 192, 336, 720\}$.
\textit{Short-term Forecasting}: The prediction horizons are comparatively short, ranging in $[6, 48]$. Correspondingly, the input lengths are set to be twice the size of the prediction horizons.

\subsection{Main Results}
\subsubsection{Long-term Forecasting}
\cref{tab::long-term} presents the quantitative results of our method and other state-of-the-art methods on long-term forecasting benchmarks.
Our TS-HFTA consistently surpasses all baselines.

These results underscore the robustness and efficiency of our TS-HFTA in capturing long-term dependencies and trends in multivariate time-series data.

Specifically, on the ETTm2 dataset, TS-HFTA achieves an MAE of 0.301, representing a significant 6.2\% improvement compared to CALF's MAE of 0.321. This improvement indicates that TS-HFTA is better equipped to minimize errors in long-horizon predictions, even for challenging datasets with complex temporal patterns. On the Electricity dataset, TS-HFTA records an MSE of 0.164, which is 6.3\% lower than CALF's 0.175. This result further highlights the model's capacity to provide more accurate predictions while reducing variance in its estimates.

Beyond these datasets, TS-HFTA also achieves competitive performance on the Weather and Traffic datasets. For example, in the Weather dataset, which contains highly seasonal and dynamic patterns, TS-HFTA's ability to model long-term dependencies results in lower prediction errors compared to both TimeLLM \cite{jin2023time} and CALF \cite{liu2024taming}. Similarly, in the Electricity dataset, where the relationships among time-series variables are intricate and interdependent, TS-HFTA outperforms competitors by effectively leveraging its architecture to capture these interactions over extended periods. 

It's noted that \textbf{PAttn} \cite{tan2024language} and \textbf{LTrsf} \cite{tan2024language} do not exhibit high performance when evaluated under a uniform and comparable setting. Notably, the sequence modeling capabilities of LLMs are significantly weakened when the input sequence length becomes excessively large. This observation aligns with insights from the recent NeurIPS 2024 workshop by Christoph Bergmeir, who emphasized that input window length is a critical hyperparameter. He further noted that restricting input lengths to smaller values often favors more complex models over simpler ones \footnote{\url{https://cbergmeir.com/talks/neurips2024/}}.

\input{tab1}

The superior performance of TS-HFTA can be attributed to its design, which emphasizes hierarchical alignment strategies. Unlike TimeLLM \cite{jin2023time}, which uses sequential structure without explicit alignment strategy, and CALF \cite{liu2024taming}, which employs naive cross-attention mechanisms, TS-HFTA adopts a prompt-enhanced cross-attention mechanism and a dynamic adaptive gating approach to activate the sequence modeling capabilities of large language models. This virtual text generation allows TS-HFTA to integrate semantics derived from compressed word embeddings while preserving the numerical characteristics of the time-series data. Consistent with recent studies \cite{jiang2024empowering, jin2024position}, our findings support the conclusion that LLMs hold significant potential for advancing time-series analysis.

\subsubsection{Short-term Forecasting}
\cref{tab::short-term} presents the quantitative results of our method and other state-of-the-art methods on short-term forecasting benchmarks.
TS-HFTA achieves the best performance, leading with the lowest SMAPE (11.651), MASE (1.563), and OWA (0.837) across average intervals, as shown in \cref{tab::short-term}, marking a significant 3.5\% improvement over CALF. These results demonstrate TS-HFTA's ability to deliver accurate short-term predictions, further highlighting its robustness in handling diverse temporal patterns.

\input{tab2}

Overall, LLMs-based methods \cite{liu2024taming,jin2023time,zhou2023one} consistently outperform traditional architectures such as MLP-based models \cite{zeng2023dlinear,das2023tide}, CNNs \cite{BaiTCN2018,wang2022micn,Wu2022TimesNetT2}, and Transformer-based models \cite{nie2022pathtst,liu2023itransformer,zhang2022crossformer,woo2022etsformer,zhou2022fedformer,wu2021autoformer}. The superiority of LLMs-based approaches lies in their enhanced ability to capture rich contextual information and adapt effectively to a wide range of tasks, enabling them to generalize across different datasets and temporal granularities.
The M4 \cite{M4team2018dataset} dataset, which encompasses multiple frequencies (yearly, quarterly, monthly, etc.), presents significant challenges for traditional architectures due to their limitations in capturing patterns specific to varying periodicities. Traditional models often struggle to adapt their feature extraction mechanisms across diverse temporal granularities, leading to suboptimal performance when faced with datasets containing mixed-frequency components. TS-HFTA's remarkable performance on this dataset demonstrates its effectiveness in modeling diverse seasonal and trend components, which are prevalent in marketing and economic data. The ability to generalize across various frequencies underscores TS-HFTA's flexibility and adaptability in real-world applications.

\subsubsection{Visualization} 
\begin{figure}[ht]
    \centering
    \begin{minipage}[b]{0.4\textwidth}
    \centering
        \includegraphics[width=\textwidth]{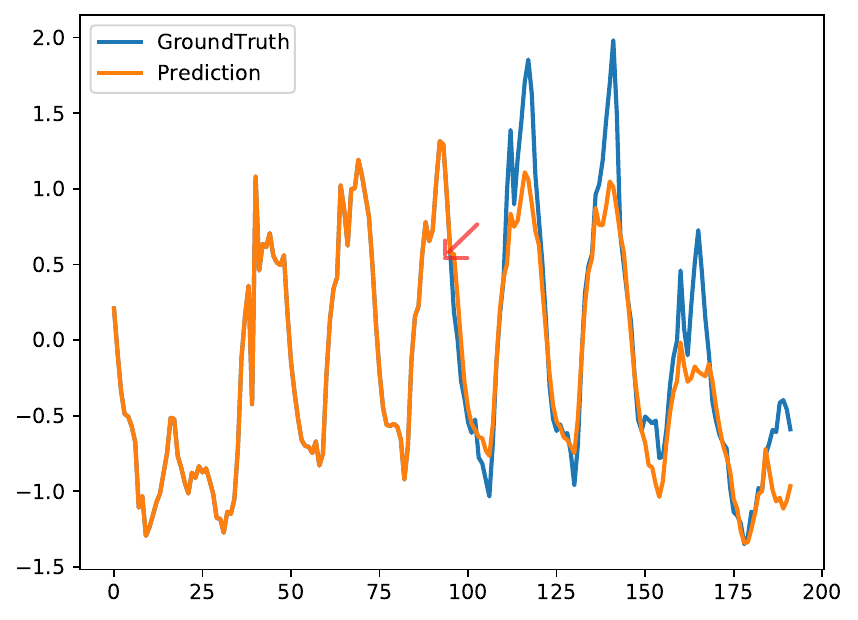}
        \caption*{\makebox[1.1\linewidth][l]{\hspace{1cm}(1) Electricity Case: 96 $\to$ 96}}
        \label{fig:sub1}
    \end{minipage}
    \hspace{0.02\textwidth}
    \begin{minipage}[b]{0.39\textwidth}
    \centering
        \includegraphics[width=\textwidth]{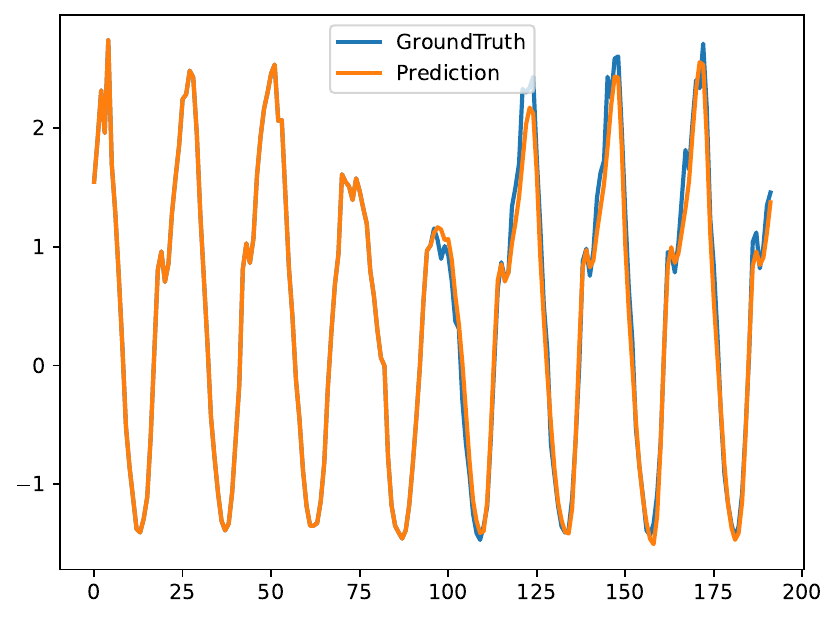}
        \caption*{\makebox[1.1\linewidth][l]{\hspace{1cm}(3) Traffic Case: 96 $\to$ 96}}
        \label{fig:sub3}
    \end{minipage}

    \vspace{0.5em} 
    \begin{minipage}[b]{0.4\textwidth}
        \includegraphics[width=\textwidth]{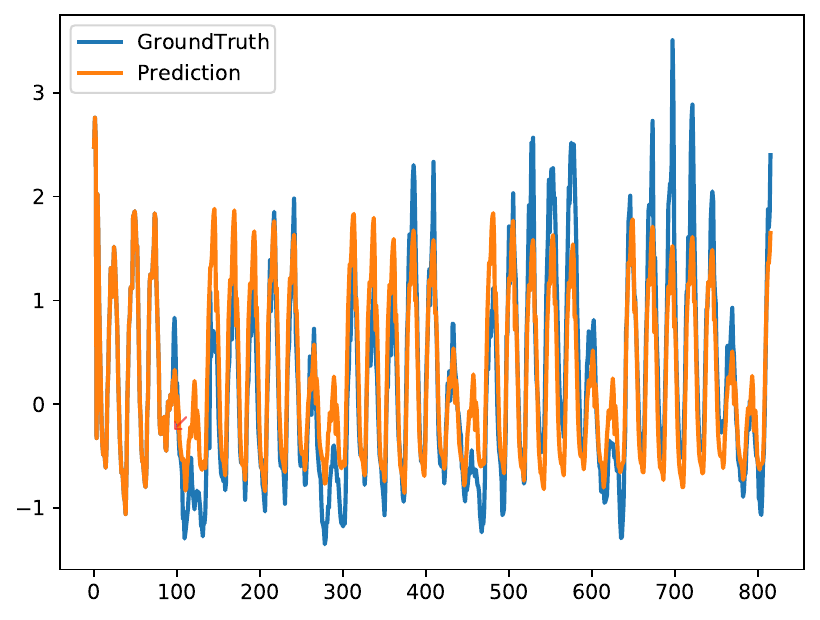}
        \centering
        \caption*{\makebox[1.1\linewidth][l]{\hspace{1cm}(2) Electricity Case: 96 $\to$720}}
        \label{fig:sub2}
    \end{minipage}
    \hspace{0.02\textwidth}
    \begin{minipage}[b]{0.4\textwidth}
    \centering
        \includegraphics[width=\textwidth]{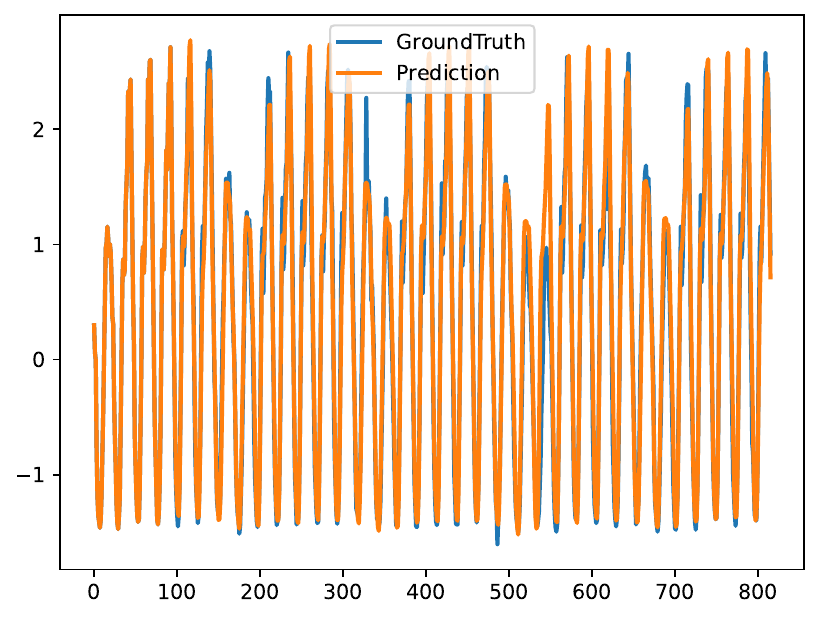}
        \caption*{\makebox[1.1\linewidth][l]{\hspace{1cm}(4) Traffic Case: 96 $\to$ 720}}
        \label{fig:sub4}
    \end{minipage}

    \caption{Long-term forecasting visualization cases for Electricity and Traffic. The blue lines represent the actual values, while the orange lines indicate the model's predictions.}
    \label{fig:analysis}
\end{figure}

\cref{fig:analysis} presents the long-term forecasting results for the Electricity and Traffic datasets under the input-96-predict-96 and input-96-predict-720 settings. Our TS-HFTA demonstrates a strong alignment between the predictions and the ground truth, especially on the Traffic dataset, where the predictions closely match the actual values. This superior performance can be attributed to the strong periodicity and regularity of traffic patterns, such as daily and weekly cycles, which the model effectively captures. 
Conversely, the Electricity dataset is characterized by higher variability and complexity. It is affected by various factors including weather, industrial activities, and unexpected events. As a result, there are occasional deviations between the predictions and the ground truth. These differences emphasize the challenges presented by noisier and less predictable data.

\subsection{Ablation Studies}  
\subsubsection{Ablation on Different Stage Alignment}

To gain a better understanding of the contribution of each component in our TS - HFTA, we carried out ablation studies on the M4 dataset. These studies isolate the effect of individual modules, offering insights into their specific roles and their synergy in achieving state-of-the-art performance. Table~\ref{tab::ablation} summarizes the results on the M4 dataset, with SMAPE, MASE, and OWA as key metrics. Lower values indicate better performance.  
\input{tab3}

\textbf{Impact of $\mathcal{L}_{DAG}$:}  
The Dynamic Adaptive Gating module ($\mathcal{L}_{DAG}$) reduces SMAPE from 11.956 to 11.821, demonstrating its pivotal role in adaptively generating the paired virtual texture token. By generating modality-specific virtual tokens while preserving numerical properties, $\mathcal{L}_{DAG}$ effectively enhances the sequence modeling capabilities, especially for large input spaces.

\textbf{Impact of $\mathcal{L}_{feature}$:}  
Adding the Intermediate Layer Alignment Feature Loss ($\mathcal{L}_{feature}$) leads to reductions in MASE and OWA, highlighting its effectiveness in stabilizing intermediate representations and refining temporal coherence. However, it results in a slight increase in SMAPE, likely due to localized trade-offs in optimizing specific temporal or feature patterns. Despite this, $\mathcal{L}_{feature}$ ensures robust multivariate correlation handling and contributes significantly to overall model performance.

\textbf{Impact of $\mathcal{L}_{ot}$:}  
Incorporating the Output Layer Alignment Optimal Transport Loss ($\mathcal{L}_{ot}$) further reduces SMAPE to 11.651. This module ensures precise alignment of output distributions, refining predictions during the final stage and contributing to overall numerical stability.

\textbf{Overall Performance:}  
The complete model, integrating all three alignments, achieves state-of-the-art performance across metrics, with SMAPE (11.651), MASE (1.563), and OWA (0.837). This confirms the necessity and complementarity of $\mathcal{L}_{DAG}$, $\mathcal{L}_{feature}$, and $\mathcal{L}_{ot}$ in advancing time-series forecasting. Each alignment contributes differently across stages. $\mathcal{L}_{DAG}$ has the greatest influence at the input stage, driving adaptive virtual texture tokens. $\mathcal{L}_{feature}$ stabilizes intermediate layers by maintaining feature integrity, while $\mathcal{L}_{ot}$ focuses on refining final predictions, ensuring output precision.

\begin{figure*}[!ht]
    \centering
\includegraphics[width=0.99\textwidth]{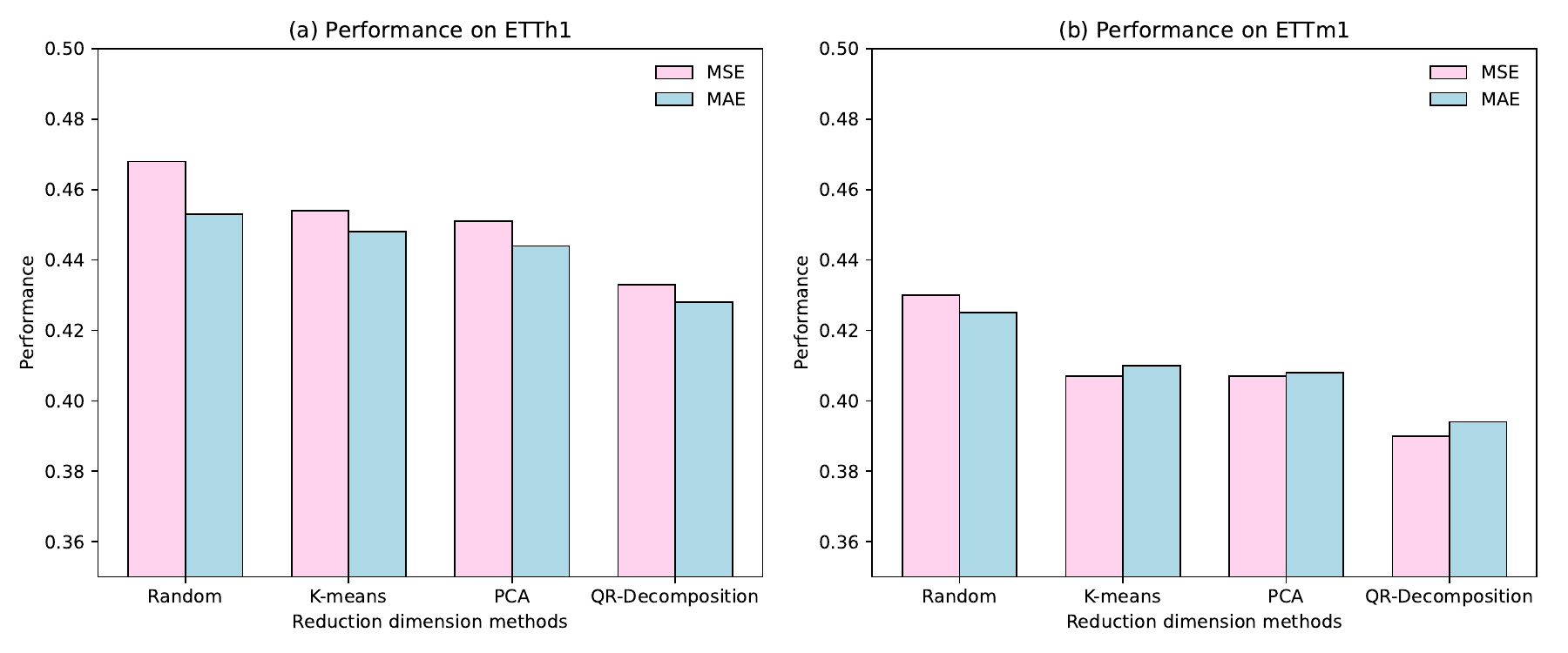}
    \caption{Ablation on different reduction methods on (a) ETTh1 and (b) ETTm1 datasets.}
    \label{fig:reduction}
\end{figure*}

\subsubsection{ Ablation on Different Reduction Methods} 
We compared QR decomposition \footnote{\url{https://numpy.org/doc/stable/reference/generated/numpy.linalg.qr.html}}, PCA \footnote{\url{https://scikit-learn.org/stable/modules/generated/sklearn.decomposition.PCA.html}}, K-means clustering \footnote{\url{https://scikit-learn.org/stable/modules/clustering.html\#k-means}}, and random selection for vocabulary compression, using the same model accuracy metric. The vocabulary size for each method is reduced to the rank of the matrix from QR decomposition.  
The \cref{fig:reduction} shows that QR decomposition achieved the highest performance. \textbf{QR decomposition's key advantage lies in maintaining orthogonality, ensuring that the prototype vectors can span the semantic space of large language models}. This preservation of semantic structure gives it an edge over PCA  and K-means, which, while effective, lose critical semantic relationships or are sensitive to initialization. Random selection performed the worst, highlighting the risks of unstructured compression. This may be the reason that in high-dimensional spaces, not all vectors are pairwise orthogonal, and this is due to semantic relationships. 

\subsubsection{Hyper-parameter Study}
\begin{figure}[ht]
    \centering
    \begin{minipage}[b]{0.3\textwidth}
        \centering
        \includegraphics[width=\textwidth]{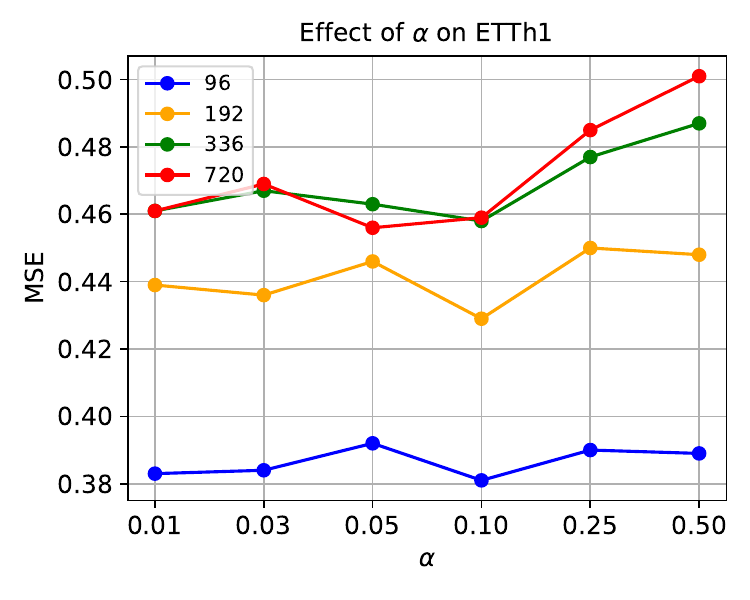}
        \caption*{\makebox[1.1\linewidth][l]{\hspace{1cm} (1) Sensitivity of $\alpha$}}
        \label{fig:sub1}
    \end{minipage}
    \hspace{0.02\textwidth}
    \begin{minipage}[b]{0.3\textwidth}
        \centering
        \includegraphics[width=\textwidth]{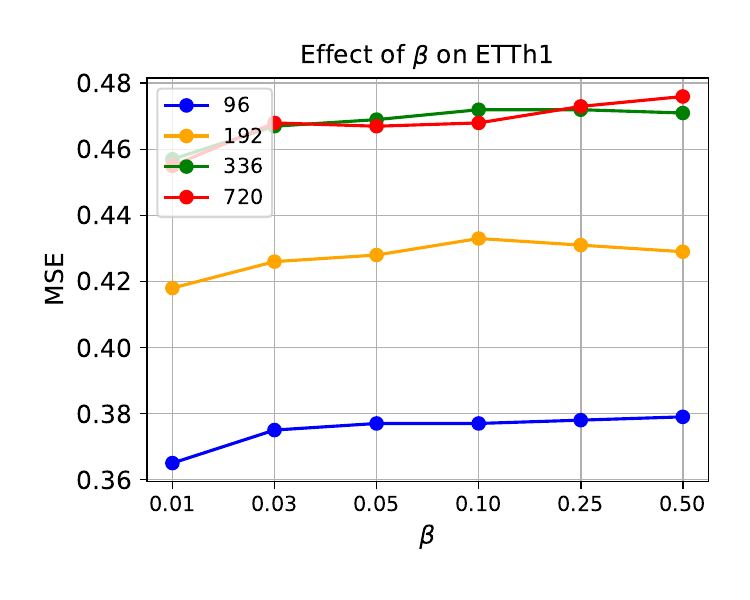}
        \caption*{\makebox[1.1\linewidth][l]{\hspace{1cm}(2) Sensitivity of $\beta$}} 
        \label{fig:sub2}
    \end{minipage}
    \hspace{0.02\textwidth}
    \begin{minipage}[b]{0.3\textwidth}
        \centering
        \includegraphics[width=\textwidth]{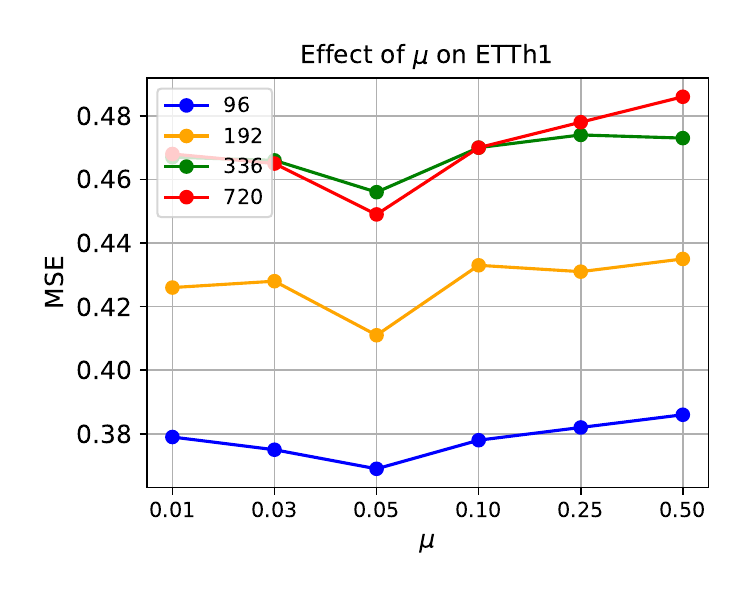}
        \caption*{\makebox[1.1\linewidth][l]{\hspace{1cm}(3) Sensitivity of $\mu$}}
        \label{fig:sub3}
    \end{minipage}

    \caption{Hyper-parameter sensitivity analysis for the ETTh1 dataset. Each subfigure presents the effect of varying $\alpha$, $\beta$, and $\mu$ values on the model’s performance.}
    \label{fig:hyper-analysis}
\end{figure}

We conducted a sensitivity analysis by varying the hyper-parameters $\alpha \in [0.01, 0.5]$, $\beta \in [0.01, 0.5]$, and $\mu \in [0.01, 0.5]$ on the ETTh1 dataset. The results are shown in \cref{fig:hyper-analysis}, the MSE remains relatively stable within these ranges. The best performance (MSE: 0.423) is achieved when $\alpha = 0.1$, $\beta = 0.01$, and $\mu = 0.05$. Although further fine-tuning may lead to minor improvements, our primary goal is to validate the robustness and effectiveness of the proposed method, rather than optimizing hyper-parameters for each dataset. Therefore, we employed a uniform set of hyper-parameters across all datasets instead of re-tuning for each one. The results demonstrate that our approach performs reliably across various settings, underscoring its practicality for diverse time-series forecasting tasks.

\input{05_4_parameter}

%% file: 05_1_baselines.tex
We introduce the baseline models selected for comparison in the following section:

\begin{itemize}[noitemsep,topsep=0pt,parsep=0pt,partopsep=0pt]
    \item \textbf{CALF} \cite{liu2024taming}: CALF employs a cross-modal matching module with a two-stream architecture to mitigate modality gaps and align features.
    \item \textbf{TimeLLM}~\citep{jin2023time}: TimeLLM reprograms time-series tokens using multi-head attention and fine-tunes pre-trained LLMs with prefix prompting for time-series analysis.
    \item \textbf{GPT4TS}~\cite{zhou2023one}: GPT4TS represents time-series data as patched tokens and fine-tunes GPT2~\cite{radford2019language} for various time-series tasks.
    \item \textbf{PatchTST}~\cite{nie2022pathtst}: PatchTST utilizes a Transformer-based model that segments data into patches and employs a channel-independent design to improve forecasting efficiency and performance.
    \item \textbf{iTransformer}~\cite{liu2023itransformer}: iTransformer captures multivariate correlations by applying attention and feed-forward networks to inverted time-series dimensions.
    \item \textbf{Crossformer}~\cite{zhang2022crossformer}: Crossformer extracts segment-wise representations channel by channel and applies a two-stage attention mechanism to capture cross-temporal and cross-dimensional dependencies.
    \item \textbf{FEDformer}~\cite{zhou2022fedformer}: FEDformer incorporates seasonal-trend decomposition and frequency-domain information into Transformers to improve efficiency and accuracy in time-series forecasting.
    \item \textbf{Autoformer}~\cite{wu2021autoformer}: Autoformer employs a decomposition architecture and Auto-Correlation mechanisms for efficient and accurate long-term forecasting.
    \item \textbf{ETSformer}~\cite{woo2022etsformer}: ETSformer integrates exponential smoothing into attention mechanisms, combining exponential smoothing attention and frequency attention for time-series forecasting.
    \item \textbf{PAttn} \cite{tan2024language} and \textbf{LTrsf} \cite{tan2024language}: These models question the effectiveness of LLMs in time-series analysis. Evaluations with a standardized lookback window of 96 and the "Drop Last" option disabled during testing \cite{qiu2024tfb} reveal their potential limitations.
    \item \textbf{MICN}~\cite{wang2022micn}: MICN employs multi-scale convolutions to capture temporal dependencies at different resolutions and models complex feature interactions for accurate forecasting.
    \item \textbf{TimesNet}~\cite{Wu2022TimesNetT2}: TimesNet transforms 1D time-series data into 2D representations, using TimesBlock with an inception module to capture intra- and inter-period relations for diverse tasks.
    \item \textbf{Dlinear}~\cite{zeng2023dlinear}: Dlinear models time-series data by decomposing and separately modeling trend and seasonal components using linear layers.
    \item \textbf{TiDE}~\cite{das2023tide}: TiDE combines temporal convolutional networks and recurrent neural networks to capture both short-term patterns and long-term dependencies effectively.
    \item \textbf{N\-BEATS} \cite{oreshkin2019nbeats}: N\-BEATS models trend and seasonality components using backward and forward residual links for interpretable time-series forecasting.
    \item \textbf{N\-HiTS} \cite{challu2022nhits}: N\-HiTS dynamically adjusts hierarchical structures to refine predictions, effectively handling multiple temporal resolutions.
\end{itemize}

%% file: 05_3_dataset.tex
\begin{table}[htbp]
  \caption{{Dataset statistics are from \cite{Wu2022TimesNetT2}. The dimension indicates the number of time series (i.e., channels), and the dataset size is organized in (training, validation, testing).}} \label{tab:dataset}
  \centering
   \resizebox{0.99\columnwidth}{!}{
  \begin{threeparttable}
  \begin{small}
  \renewcommand{\multirowsetup}{\centering}
  \setlength{\tabcolsep}{3.8pt}
  \begin{tabular}{c|l|c|c|c|c|c}
    \toprule
    Task & Dataset & Dim. & Series Length & Dataset Size &Frequency  &\scalebox{1.0}{\revision{Domain}} \\
    \toprule
     & ETTm1 & 7 & \scalebox{0.8}{\{96, 192, 336, 720\}} & (34465, 11521, 11521)  & 15 min &\scalebox{1.0}{Temperature}\\
    \cmidrule{2-7}
    & ETTm2 & 7 & \scalebox{0.8}{\{96, 192, 336, 720\}} & (34465, 11521, 11521)  & 15 min &\scalebox{1.0}{Temperature}\\
     \cmidrule{2-7}
     Long-term& ETTh1 & 7 & \scalebox{0.8}{\{96, 192, 336, 720\}} & (8545, 2881, 2881) & 1 hour &\scalebox{1.0}{Temperature} \\
     \cmidrule{2-7}
    Forecasting  &ETTh2 & 7 & \scalebox{0.8}{\{96, 192, 336, 720\}} & (8545, 2881, 2881) & 1 hour &\scalebox{1.0}{Temperature} \\ 
     \cmidrule{2-7}
     &\revision{Electricity} & \revision{321} & \revision{\scalebox{0.8}{\{96, 192, 336, 720\}}} & \revision{(18317, 2633, 5261)} & \revision{1 hour} &\scalebox{1.0}{\revision{Electricity}} \\ 
     \cmidrule{2-7}
     &\revision{Traffic} & \revision{862} & \revision{\scalebox{0.8}{\{96, 192, 336, 720\}}} & \revision{(12185, 1757, 3509)} & \revision{1 hour} &\scalebox{1.0}{\revision{Transportation}} \\ 
     \cmidrule{2-7}
     &\revision{Weather} & \revision{21} & \revision{\scalebox{0.8}{\{96, 192, 336, 720\}}} & \revision{(36792, 5271, 10540)} & \revision{10 min} &\scalebox{1.0}{\revision{Weather}} \\
    \midrule
    & M4-Yearly & 1 & 6 & (23000, 0, 23000) &Yearly  & \scalebox{1.0}{Demographic} \\
    \cmidrule{2-7}
     Short-term  & M4-Quarterly & 1 & 8 & (24000, 0, 24000) &Quarterly  & \scalebox{1.0}{Finance} \\
    \cmidrule{2-7}
    Forecasting & M4-Monthly & 1 & 18 & (48000, 0, 48000) & Monthly & \scalebox{1.0}{Industry} \\
     \cmidrule{2-7}
     & M4-Weakly & 1 & 13 & (359, 0, 359) & Weakly & \scalebox{1.0}{Macro} \\
     \cmidrule{2-7}
     & M4-Daily & 1 & 14 & (4227, 0, 4227) &Daily  & \scalebox{1.0}{Micro} \\
     \cmidrule{2-7}
     & M4-Hourly & 1 &48 & (414, 0, 414) & Hourly  & \scalebox{1.0}{Other} \\
    \bottomrule
    \end{tabular}
    \end{small}
  \end{threeparttable}
  }
\end{table}

%% file: 05_2_metrics.tex
The calculations of these metrics are as follows:
\begin{align*} \label{equ:metrics}
    \text{MSE} &= \frac{1}{H}\sum_{h=1}^T (\mathbf{Y}_{h} - \Hat{\mathbf{Y}}_{h})^2,
    &
    \text{MAE} &= \frac{1}{H}\sum_{h=1}^H|\mathbf{Y}_{h} - \Hat{\mathbf{Y}}_{h}|,\\
    \text{SMAPE} &= \frac{200}{H} \sum_{h=1}^H \frac{|\mathbf{Y}_{h} - \Hat{\mathbf{Y}}_{h}|}{|\mathbf{Y}_{h}| + |\Hat{\mathbf{Y}}_{h}|},
    &
    \text{MAPE} &= \frac{100}{H} \sum_{h=1}^H \frac{|\mathbf{Y}_{h} - \Hat{\mathbf{Y}}_{h}|}{|\mathbf{Y}_{h}|}, \\
    \text{MASE} &= \frac{1}{H} \sum_{h=1}^H \frac{|\mathbf{Y}_{h} - \Hat{\mathbf{Y}}_{h}|}{\frac{1}{H-s}\sum_{j=s+1}^{H}|\mathbf{Y}_j - \mathbf{Y}_{j-s}|},
    &
    \text{OWA} &= \frac{1}{2} \left[ \frac{\text{SMAPE}}{\text{SMAPE}_{\textrm{Naïve2}}}  + \frac{\text{MASE}}{\text{MASE}_{\textrm{Naïve2}}}  \right].
\end{align*}
Here, $s$ represents the periodicity of the time-series data, and $H$ denotes the number of data points (i.e., the prediction horizon in our case). $\mathbf{Y}_{h}$ and $\Hat{\mathbf{Y}}_{h}$ correspond to the ground truth and predicted values at the $h$-th step, where $h \in \{1, \cdots, H\}$.

%% file: tab1.tex
\begin{table}[!ht]
\centering
\scriptsize
\caption{Multivariate long-term forecasting results with input length $T=96$. Results are averaged over 4 prediction lengths $H \in \{96, 192, 336, 720\}$. A lower value indicates better performance, and this applies to all subsequent tables.}
\begin{adjustwidth}{-0.1cm}{0cm}
\begin{tabular}{c|c|ccccccc|c}
\toprule
\textbf{Method} & \textbf{Metric} $\downarrow$ & \textbf{ETTm1} & \textbf{ETTm2} & \textbf{ETTh1} & \textbf{ETTh2 }& \textbf{Weather} & \textbf{Electricity} & \textbf{Traffic} & \textbf{\textbf{1st Count}} \\
\midrule
\multirow{2}{*}{TiDE \cite{das2023tide}} & MSE  & 0.412 & 0.289 & 0.445 & 0.611 & 0.271 & 0.251 & 0.760 & \multirow{2}{*}{0} \\
                       & MAE &  0.406 & 0.326 & 0.432 & 0.550 & 0.320 & 0.344 & 0.473 & \\
\midrule
\multirow{2}{*}{DLinear \cite{zeng2023dlinear}} & MSE  & 0.403 & 0.350 & 0.456 & 0.559 & 0.265 & 0.212 & 0.625 & \multirow{2}{*}{0} \\
                       & MAE  & 0.407 & 0.401 & 0.452 & 0.515 & 0.317 & 0.300 & 0.383 & \\  
\midrule
\multirow{2}{*}{MICN \cite{wang2022micn}} & MSE  & 0.392 & 0.328 & 0.558 & 0.587 & \best{0.242} & 0.186 & 0.541 & \multirow{2}{*}{1} \\
                       & MAE  & 0.413 & 0.382 & 0.535 & 0.525 & 0.299 & 0.294 & 0.315 & \\    
\midrule
\multirow{2}{*}{TimesNet \cite{Wu2022TimesNetT2}} & MSE & 0.400 & 0.291 & 0.458 & 0.414 & 0.259 & 0.192 & 0.620 & \multirow{2}{*}{0} \\
                       & MAE & 0.406 & 0.333 & 0.450 & 0.427 & 0.287 & 0.295 & 0.336 & \\ 
\midrule
\multirow{2}{*}{FEDformer \cite{zhou2022fedformer}} & MSE  & 0.448 & 0.305 & 0.440 & 0.437 & 0.309 & 0.214 & 0.610 & \multirow{2}{*}{0} \\
                       & MAE & 0.452 & 0.349 & 0.460 & 0.449 & 0.360 & 0.327 & 0.376 & \\ 
                       \midrule
\multirow{2}{*}{Crossformer \cite{zhang2022crossformer}} & MSE  & 0.502 & 1.216 & 0.620 & 0.942 & 0.259& 0.244 & 0.550& \multirow{2}{*}{0} \\
                       & MAE & 0.502 & 0.707 & 0.572 & 0.684 & 0.315 & 0.334 & 0.304 & \\ 
\midrule
\multirow{2}{*}{iTransformer \cite{liu2023itransformer}} & MSE  & 0.407 & 0.291 & 0.455 & 0.381 & 0.257& 0.178 & \second{0.428}& \multirow{2}{*}{0} \\
                       & MAE & 0.411& 0.335 & 0.448 & 0.405 & 0.279 & 0.270 & \second{0.282} & \\ 
                       \midrule
\multirow{2}{*}{PatchTST \cite{nie2022pathtst}} & MSE & \best{0.381} & 0.285 & 0.450 & 0.366 & 0.258& 0.216 & 0.555& \multirow{2}{*}{1} \\
                       & MAE  & 0.395 & 0.327 & 0.441 & 0.394 & 0.280 & 0.304 & 0.361 & \\      
\midrule
\multirow{2}{*}{GPT4TS \cite{zhou2023one}} & MSE & 0.389 & 0.285 & 0.447 & 0.381 & 0.264 & 0.205 & 0.488 & \multirow{2}{*}{0} \\
                       & MAE  & 0.397 & 0.331 & 0.436 & 0.408 & 0.284 & 0.290 & 0.317 & \\ 
\midrule
\multirow{2}{*}{TimeLLM \cite{jin2023time}}  & MSE & 0.410 & 0.296 & 0.460 & 0.389 & 0.274 & 0.223 & 0.541 & \multirow{2}{*}{0} \\
                       & MAE  & 0.409 & 0.340 & 0.449 & 0.408 & 0.290 & 0.309 & 0.358 & \\ 
\midrule
\multirow{2}{*}{PAttn \cite{tan2024language}}  & MSE  & 0.390 & \second{0.281} & 0.449 & 0.369 & 0.261 & 0.209 & 0.562 & \multirow{2}{*}{0} \\
                       & MAE  & 0.386 & 0.320 & 0.428 & 0.392 & 0.276 & 0.282 & 0.331 & \\ 
\midrule
\multirow{2}{*}{LTrsf \cite{tan2024language}}  & MSE  & 0.400 & 0.282 & 0.446 & 0.374 & 0.262 & 0.201 & 0.518 & \multirow{2}{*}{0} \\
                       & MAE & 0.392 & 0.322 & 0.433 & 0.397 & 0.275 & 0.275 & 0.312 & \\ 

\midrule
\multirow{2}{*}{CALF \cite{liu2024taming}} & MSE  & 0.395 & \second{0.281} & \second{0.432} & \best{0.349} & 0.250 & \second{0.175} & 0.439 & \multirow{2}{*}{\second{2}} \\
                       & MAE  & \second{0.390} & \second{0.321} & \second{0.428} & \second{0.382} & \second{0.274} & \second{0.265} & \best{0.281} & \\ 
                       \midrule
\multirow{2}{*}{TS-HFTA (\textbf{Ours})} & MSE  &\second{0.383} & \best{0.276} & 
\best{0.423} & \second{0.351} & \second{0.248}& \best{0.164} & \best{0.425}& \multirow{2}{*}{\best{10}} \\
                       & MAE  & \best{0.373} & \best{0.301} & \best{0.405} & \best{0.361} & \best{0.260} & \best
                       {0.254} & 0.285 & \\
\bottomrule
\end{tabular}
\end{adjustwidth}
\label{tab::long-term}
\end{table}

%% file: tab2.tex
\begin{table}[!ht]
\centering
\caption{Short-term forecasting results on M4 dataset. Input lengths are $[12, 96]$ and prediction are $[6, 48]$. The rows provided are weighted averaged from all datasets under different sampling intervals.}
\scalebox{0.85}{
\begin{tabular}{c|c|ccc|c} 
\toprule
\textbf{Method} &\textbf{Source} & \textbf{SMAPE} $\downarrow$ & \textbf{MASE} $\downarrow$ & \textbf{OWA} $\downarrow$ & \textbf{1st Count} \\
\midrule
TCN \cite{BaiTCN2018} & Arxiv'2018&13.961 & 1.945 & 1.023 & 0 \\
TimesNet \cite{Wu2022TimesNetT2} & ICLR'2023 &11.829 & 1.585 & 0.851 & 0 \\
\midrule
DLinear \cite{zeng2023dlinear} & AAAI'2023&13.639 & 2.095 & 1.051 & 0 \\
N-BEATS \cite{oreshkin2019nbeats} & ICLR'2020 &11.851 & 1.599 & 0.855 & 0 \\
N-HiTS \cite{challu2022nhits} & AAAI'2023 &11.927 & 1.613 & 0.861 & 0 \\
\midrule
Autoformer \cite{wu2021autoformer}  & NeurIPS'2021 & 12.909 & 1.771 & 0.939 & 0 \\
FEDformer \cite{zhou2022fedformer} & ICML 2022 &12.840 & 1.701 & 0.918 & 0 \\
ETSformer \cite{woo2022etsformer}  & ICLR'2023 &14.718 & 2.408 & 1.172 & 0 \\
PatchTST \cite{nie2022pathtst} & ICLR'2023 &12.059 & 1.623 & 0.869 & 0 \\
\midrule
GPT4TS \cite{zhou2023one} & NeurIPS'2023 &11.991 & 1.600 & 0.861 & 0 \\
TimeLLM \cite{jin2023time} & ICLR'2024 & 11.983 & 1.595 & 0.859 & 0 \\
CALF \cite{liu2024taming} & Arxiv'2024 &\second{11.765} & \second{1.567} & \second{0.844} & 0 \\
\midrule
TS-HFTA (\textbf{Ours})  &  Arxiv'2024 &\best{11.651} & \best{1.563} & \best{0.837} & \best{3} \\
\bottomrule
\end{tabular}
}
\label{tab::short-term}
\end{table}

%% file: tab3.tex
\begin{table}[htb]
\setlength{\tabcolsep}{5pt} 
\renewcommand{\arraystretch}{1.2} 
\centering
\caption{Ablation study on the M4 dataset using average interval metrics. }
\scalebox{0.85}{
\begin{tabular}{c|ccc|ccc} 
\toprule
\multirow{2}{*}{\textbf{Model Configuration}} & 
\multicolumn{3}{c|}{\textbf{Components Enabled}} & 
\multicolumn{3}{c}{\textbf{M4 Metrics}} \\
\cmidrule(lr){2-4} \cmidrule(lr){5-7}
& $\mathcal{L}_{DAG}$ & $\mathcal{L}_{feature}$ & $\mathcal{L}_{ot}$ & \textbf{SMAPE} $\downarrow$ & \textbf{MASE} $\downarrow$ & \textbf{OWA} $\downarrow$ \\
\midrule
Task (Baseline) & $-$ & $-$ & $-$ & 11.956 & 1.678 & 0.892 \\
DAG Only & \checkmark & $-$ & $-$ & \second{11.821} & 1.621 & 0.881 \\
DAG + Feature & \checkmark & \checkmark & $-$ & 11.844 & \second{1.567} & \second{0.864} \\
DAG + Feature + OT (\textbf{Ours}) & \checkmark & \checkmark & \checkmark & \best{11.651} & \best{1.563} & \best{0.837} \\
\bottomrule
\end{tabular}
}
\label{tab::ablation}
\end{table}

%% file: 05_4_parameter.tex
\subsubsection{Complexity Analysis and Model Efficiency}
In this section, we present a comprehensive evaluation of the computational complexity and efficiency of the proposed TS-HFTA. We compare it against various baselines, highlighting both the theoretical complexity and practical efficiency in terms of floating-point operations (FLOPs), parameter count, training speed, and prediction performance.

\renewcommand{\arraystretch}{1.5}
\begin{table*}[htb]
\setlength{\tabcolsep}{7pt}
\centering
\caption{Per-layer theoretical computational complexity of Transformer-based methods. Here, $T$ and $H$ represent the lengths of the input and prediction sequences, respectively, $C$ denotes the number of channels, and $p$ denotes the patch length in patch-based methods.}
\scalebox{0.85}{
\begin{tabular}{c|c|c}
\toprule
\textbf{Method} & \textbf{Encoder Complexity} & \textbf{Decoder Complexity} \\
\midrule
Autoformer \cite{wu2021autoformer} & $O(T \log T)$ & $O((\frac{T}{2}+H) \log (\frac{T}{2}+H))$ \\
FEDformer \cite{zhou2022fedformer} & $O(T)$ & $O(\frac{T}{2}+H)$ \\
ETSformer \cite{woo2022etsformer} & $O(T \log T)$ & $O(T \log H)$ \\
Crossformer \cite{zhang2022crossformer} & $O(\frac{C}{p^2} T^2)$ & $O(\frac{C}{p^2} H(T+H))$ \\
PatchTST \cite{nie2022pathtst} & $O((\frac{T}{p})^2)$ & - \\
iTransformer \cite{liu2023itransformer} & $O(C^2)$ & - \\
GPT4TS \cite{zhou2023one} & $O((\frac{T}{p})^2)$ & - \\
Time-LLM \cite{jin2023time} & $O((\frac{T}{p})^2)$ & - \\
CALF \cite{liu2024taming} & $O(C^2)$ & - \\
\midrule
TS-HFTA (\textbf{Ours}) & $O(C^2)$ & - \\
\bottomrule
\end{tabular}
}
\label{complexity}
\end{table*}

\textbf{Theoretical Complexity:}  \cref{complexity} shows the theoretical computational complexity per layer for various Transformer-based models. Unlike conventional Transformer-based approaches, whose computational complexity grows quadratically with the input sequence length $T$, our TS-HFTA model, inspired by \cite{liu2023itransformer}, primarily associates its complexity with the number of channels $C$. Since the temporal dimension in time-series data is typically much larger than the number of channels, the inverted Transformer architecture effectively captures inter-channel dependencies. This design choice significantly reduces the overall computational complexity compared to other models.

\begin{table}[ht]
\centering
\caption{Efficiency comparison under input-96-predict-96 setting on the ETTh1 dataset.}
\label{tab:efficiency_comparison}
\scalebox{0.85}{
\begin{tabular}{c|cccc}
\toprule
\textbf{Method} & \textbf{FLOPs} & \textbf{Parameter} & \textbf{Training (ms/iter)} & \textbf{MSE} \\
\midrule
FEDformer \cite{zhou2022fedformer}      & 38.043G & 10.536M & 260.10 & 0.376 \\
DLinear \cite{zeng2023dlinear}        & 4.150M  & 18.624K & 5.03   & 0.386 \\
TimesNet \cite{Wu2022TimesNetT2}      & 18.118G & 605.479K & 48.25 & 0.384 \\
TiDE \cite{das2023tide}          & 389.810M & 1.182M  & 34.28 & 0.479 \\
Crossformer \cite{zhang2022crossformer}   & 62.953G & 42.063M & 56.34  & 0.423 \\
PatchTST \cite{nie2022pathtst}      & 8.626G  & 3.752M  & 10.41  & 0.414 \\
iTransformer \cite{liu2023itransformer}  & 78.758M & 224.224K & 10.22 & 0.386 \\
GPT4TS \citep{zhou2023one}   & 82.33M & 42.063M & 56.34  & 0.423 \\
TimeLLM \cite{jin2023time}     & 3.4G  & 32136M  & 517.37  & \second{0.362} \\
CALF \cite{liu2024taming} & 1.3G & 18.02M & 80.22 & 0.369 \\
\midrule
TS-HFTA (\textbf{Ours})          & 1.2G  & 19.97M  & 73.67  & \best{0.356} \\
\bottomrule
\end{tabular}
}
\label{tab:model_efficiency}
\end{table}

\textbf{Model Efficiency:} This section presents a detailed comparison between the baseline models and the proposed TS-HFTA model in terms of floating-point operations (FLOPs), parameter count, training speed, and prediction performance. \cref{tab:model_efficiency} reports the results on the ETTh1 dataset, where all models use an input and prediction length of 96 with the same batch size. The proposed model consistently outperforms other Transformer-based and CNN-based models while maintaining comparable or lower computational complexity for LLM-based models and competitive training times. Although our model has relatively higher FLOPs compared to traditional architectures, the overall number of trainable parameters remains low. Additionally, LoRA \cite{hu2021lora} can further reduce the parameter count while maintaining performance.

Notably, the computational cost during training corresponds to the dual-tower structure; however, only a single-tower structure is needed during inference for time-series data, reducing the computational overhead by nearly half. Additionally, various inference optimization strategies for large language models, such as quantization, can further improve efficiency, representing promising directions for future exploration and optimization.

%% file: 06_conclusion.tex
\section{Conclusion}
In this paper, we propose \textbf{TS-HFTA}, a novel framework designed to enhance time-series forecasting by integrating LLMs through hierarchical text-free alignment. Unlike existing methods that focus on individual alignment components, TS-HFTA provides a holistic approach by aligning the input, feature, and output layers across modalities. Our dynamic adaptive gating mechanism addresses input modality mismatches by generating virtual text tokens coherently aligned with time-series data. Additionally, layer-wise contrastive learning ensures intermediate feature consistency, while optimal transport-driven output alignment reduces task-level discrepancies. Experimental results highlight the framework's superior performance across multiple benchmarks, confirming the value of integrating these mechanisms.